\documentclass[conference]{IEEEtran}
\usepackage{times}

\usepackage[numbers]{natbib}
\usepackage{multicol}
\usepackage[bookmarks=true]{hyperref}
\usepackage{amssymb}
\usepackage{amsmath}
\usepackage{subfigure}
\usepackage{epsfig}
\usepackage{algorithm}
\usepackage[noend]{algpseudocode}
\DeclareMathOperator*{\argmax}{argmax}
\usepackage{amsthm}
\usepackage{wrapfig}

\begin{document}

\title{Learning visual servo policies via planner cloning}

\author{\authorblockN{Ulrich Viereck$^{*}$, Kate Saenko$^{+}$, Robert Platt$^{*}$}
\authorblockA{$^{*}$Khoury College of Computer Sciences, Northeastern University\\ $^{+}$Department of Computer Science, Boston University}}

\maketitle

\begin{abstract}

Learning control policies for visual servoing in novel environments is an important problem. However, standard model-free policy learning methods are slow. This paper explores planner cloning: using behavior cloning to learn policies that mimic the behavior of a full-state motion planner in simulation. We propose Penalized Q Cloning (PQC), a new behavior cloning algorithm. We show that it outperforms several baselines and ablations on some challenging problems involving visual servoing in novel environments while avoiding obstacles. Finally, we demonstrate that these policies can be transferred effectively onto a real robotic platform, achieving approximately an 87\% success rate both in simulation and on a real robot.

\end{abstract}

\IEEEpeerreviewmaketitle

\section{Introduction}

Visual servoing in novel environments is an important problem. Given images produced by a camera, a visual servo control policy guides a grasped part into a desired pose relative to the environment. This problem appears in many situations: reaching, grasping, peg insertion, stacking, machine assembly tasks, etc. Whereas classical approaches to the problem~\cite{hutchinson1996tutorial,visual_servoing_classical,yoshimi1994active} typically make strong assumptions about the environment (fiducials, known object geometries, etc.), there has been a surge of interest recently in using deep learning methods to solve these problems in more unstructured settings that incorporate novel objects~\cite{zhu2018reinforcement,mahler2017learning,yan2017sim,James17,Sadeghi17,sim2real,Levine_jmlr_2016,Levine2016}. However, it is still unclear what are the best methods here. Standard model-free reinforcement learning methods such as DQN~\cite{mnih2015dqn}, AC2~\cite{mnih2016asynchronous}, PPO~\cite{schulman2017proximal}, etc. do poorly because they treat this as an unstructured model-free problem. One way to improve upon this situation is to incorporate additional knowledge via reward shaping or curriculum learning, but these approaches can require significant design effort. Most current approaches to visual servoing in unstructured domains involve some form of imitation learning or behavior cloning~\cite{zhu2018reinforcement,mahler2017learning,yan2017sim,James17,Sadeghi17,sim2real}. However, there are substantial differences between these methods and their relative performance is unclear.

This paper focuses on a class of imitation learning approaches that we call \emph{planner cloning}, where the expert is an approximately optimal motion planner that generates estimates of the optimal value function and optimal policy given full state feedback. This information is cloned onto a value function or policy over the observation space (i.e. over the space of camera images) that can be used at test time where full state feedback is unavailable. Within this simple framework, we explore several existing and new cloning techniques and ablations in order to determine which ideas perform best on the visual servo problems we consider. We focus on a particularly challenging class of visual servo tasks that involve servoing to a goal pose while avoiding obstacles in novel environments.

The main contribution of this paper is a new behavior cloning algorithm that we call \emph{Penalized Q Cloning} (PQC). It has the following characteristics:

\begin{itemize}

    \item It learns a value function using supervised value targets provided by the planner, similar to AGGREVATE~\cite{Ross2014aggrevate}.
    
    \item It incorporates a penalty into the value targets for suboptimal actions, similar to DQfD~\cite{hester2018dqfd}.
    
    \item For every experience in the replay buffer, it updates the values for all feasible actions from that state, not just the action experienced.

\end{itemize}

\begin{wrapfigure}{r}{0.25\textwidth}
  \begin{center}
    \includegraphics[width=0.25\textwidth]{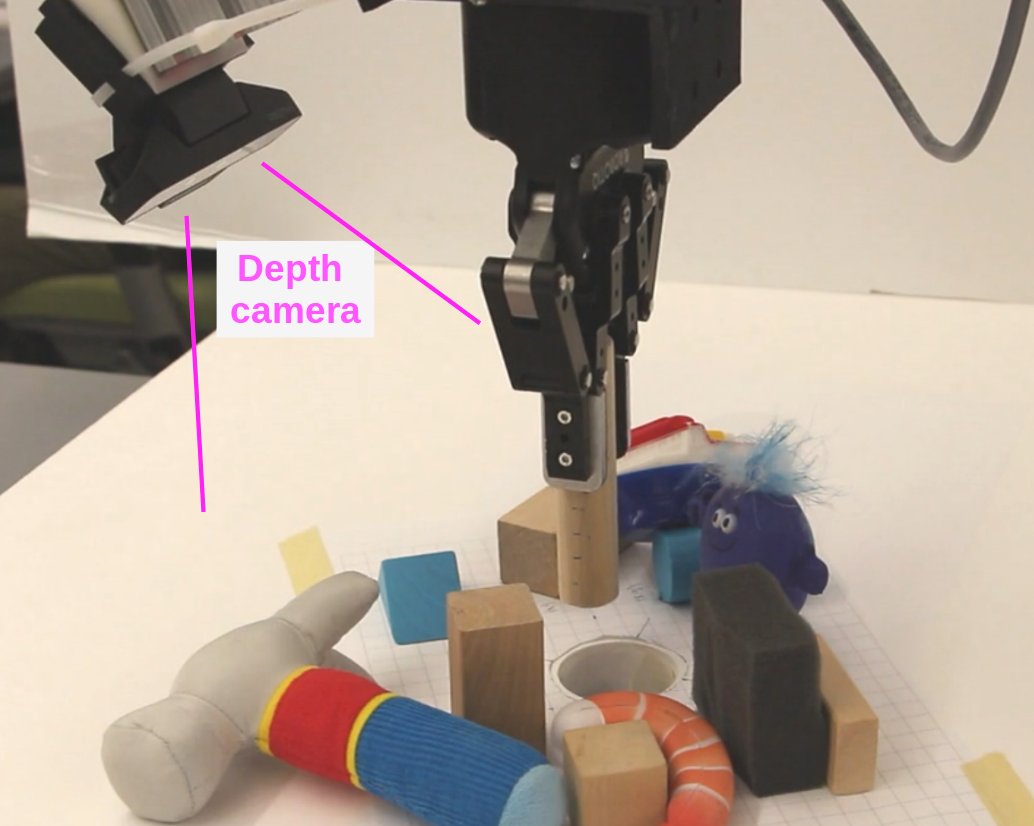}
  \end{center}
  \caption{One of the visual servoing scenarios (inserting a peg into a hole) used in our evaluations.}
  \label{fig:intro}
\end{wrapfigure}

\noindent
This algorithm differs from AGGREVATE because it incorporates the value penalties and from DQfD because it uses supervised targets rather than TD targets. We compare PQC with several baselines and algorithm ablations and show that it outperforms all these variations on two challenging visual servoing problems in novel cluttered environments, such as that shown in Figure~\ref{fig:intro}. The resulting policies achieve high success rates (approximately 87\%) on challenging visual servo tasks.

Finally, we make three additional minor contributions:
\begin{enumerate}

    \item We show that an approximately optimal motion planning algorithm can be used to produce a near optimal value function and policy that can be used for behavior cloning.

    \item We show that, in the visual servo setting at least, it is often advantageous to ``finetune'' the cloned policy using TD learning (DQN or the like) as a separate postprocessing step.
    
    \item We show that it is sometimes advantageous to use prioritized replay in the behavior cloning setting, just as it is advantageous with DQN.
    
\end{enumerate}

\subsection{Related work}

Many approaches to learning visual servo policies use standard behavior cloning where rollouts from a hand-crafted expert are used to train a policy. For example, Zhang et al. train a reaching controller to imitate a hand crafted policy that moves directly toward a target object~\cite{sim2real}. Viereck et al. do something similar using an L1 loss term~\cite{Viereck17}. Sadeghi et al. combine a policy cloning term with a TD loss term~\cite{Sadeghi17}. Yan et al. take a similar approach, combining DAGGER with a planner-based expert to solve a relatively simple visual servoing task that does not involve clutter or obstacles~\cite{yan2017sim}. In contrast to the above, this paper: 1) explores a setting where the expert is a motion planner that finds obstacle free trajectories rather than a hand crafted policy; 2) explores a variety of behavior cloning methods beyond the standard off-policy behavior cloning paradigm. 

Rather than imitating a simple hand crafted policy, Zhu et al.~\cite{zhu2018reinforcement} propose an approach to learning visuomotor tasks focused on leveraging a small number of demonstrations using on a combination of GAIL~\cite{ho2016generative} and PPO~\cite{schulman2017proximal}. They incorporate a variety of additional algorithmic pieces including curriculum learning, an asymmetric actor-critic method, and an object-centric state representation into their method. In contrast, this paper focuses on characterizing the space of cloning algorithms in a relatively simple planner cloning setting.

There are a couple of applications of imitation learning to task-level learning to point out. Nair et al.~\cite{nair2018overcoming} explore an approach to imitation learning similar to ADET~\cite{adet} where a DAGGER-like loss term is combined with a TD loss term. They demonstrate on some block stacking results in an environment with full state feedback. Mahler et al.~\cite{mahler2017learning} learn a policy for picking a sequence of objects from a bin where the actions are target grasp poses. James et al.~\cite{James17} use a variant of standard behavior cloning to clone a heuristic policy that solves a pick and place task in a non-cluttered environment.

Other work explores applications of model-free reinforcement learning (RL) to visual servoing. Well known examples are~\cite{Levine_jmlr_2016} which uses Guided Policy search to learn manipulation tasks and~\cite{Levine2016} which uses a form of Monte Carlo RL to find reach/grasp policies. One important idea in this space is that of an ``asymmetric'' actor-critic~\cite{pinto2017asymmetric} where the critic takes full state feedback while the actor takes raw observations. Our paper leverages a similar idea by doing planning in a fully observable MDP and projecting the resulting value function and policy onto a policy over observations.

There are several relevant approaches to behavior cloning that are important to mention. One of the earliest approaches~\cite{bain1999framework} generates rollouts from an expert policy that are used to train the cloned policy. DAGGER~\cite{ross2011dagger} is similar, but it combines the expert rollouts with rollouts generated by the learned policy. This results in better performance over the on-policy state distribution. Another important idea is that of cloning a value function rather than a policy, first proposed by Ross and Bagnell in the form of an algorithm called AGGREVATE~\cite{Ross2014aggrevate}. A couple of different approaches combine DQN~\cite{mnih2015dqn} with an additional term that causes a preference for expert actions. One method is DQfD~\cite{hester2018dqfd} which uses a large margin loss. The other is ADET~\cite{adet} which adds a standard cross entropy term on the policy. These above approaches are reviewed in more detail below.

\section{Background}
\label{sect:background}

\noindent
\textbf{MDPs:} We operate within the Markov decision process MDP framework. An MDP is a tuple $\mathcal{M} = (S,A,R,T,\gamma,\rho_0)$, where $S$ is a set of states, $A$ a set of actions, $R(s,a)$ is the expected reward for taking action $a$ in state $s$, $T(s,a,s')$ is the probability of transitioning to $s'$ when taking action $a$ from $s$, $\gamma$ is the temporal discount factor, and $\rho_0$ is a distribution over starting states. A stochastic policy $\pi(a | s)$ is a probability distribution over action conditioned on state. If the policy is deterministic, it can be described as a mapping, $\pi: S \mapsto A$. The value $Q^\pi(s,a)$ of a given state-action pair $(s,a)$ is the expected discounted future reward obtained starting from $(s,a)$ when following policy $\pi$. The optimal policy is the policy that takes actions so as to maximize value,
\begin{equation}
\pi^* = \argmax_{\pi} \mathbb{E}_{s \sim \rho_{\pi,\rho_0}, a \sim \pi(a|s)} \Big[ Q^\pi(s,a) \Big],
\end{equation}
where $\rho_{\pi,\rho_0}$ is the distribution over states induced by $\rho_0$ and $\pi$. The optimal action value function is $Q^*(s,a) = Q^{\pi^*}(s,a)$.

\noindent
\textbf{TD learning:} TD learning (i.e. model free reinforcement learning) calculates the optimal value function directly from trial and error experiences by minimizing the loss,
\begin{equation}
\label{eqn:q_loss}
\mathcal{L}_{TD} = \mathbb{E}_{(s,a,s',r) \sim env(\pi)} \Big[ L \big( Q(s,a), \hat{Q}(s,a) \big) \Big],
\end{equation}
where $\hat{Q}(s,a) = r + \gamma \max_{a' \in A} Q(s',a')$ is the temporal difference (TD) target, $env(\pi)$ is an empirical distribution of experiences generated by the environment under exploration policy $\pi$, and $L$ is a suitable loss function. In TD learning, the exploration policy improves during learning, approaching the optimal policy.

\noindent
\textbf{Standard Behavior Cloning:} Behavior cloning assumes the existence of an expert in the form of expert demonstrations $\mathcal{D}$, an expert policy $\pi_E$, or an expert value function $Q_E$. The expert is assumed to be good, but not necessarily optimal. The standard approach~\cite{bain1999framework} (for discrete actions) is to find a policy $\pi$ that most closely matches the behavior of the expert, i.e. that minimizes
\begin{equation}
\label{eqn:bc_loss}
\mathcal{L}_{BC} = \mathbb{E}_{s \sim \rho_{exp}} \Big[ L(\pi_E(\cdot|s), \pi(\cdot|s))\Big],
\end{equation}
where $\rho_{exp}$ is the distribution of states visited by the expert and $L$ is a suitable categorical loss function, e.g. the cross entropy loss.

\noindent
\textbf{DAGGER:} One problem with Equation~\ref{eqn:bc_loss} is that the state-action distribution over which the expectation is evaluated is different than that which the learned policy ultimately experiences. This mismatch is often corrected using DAGGER~\cite{ross2011dagger}, which gradually shifts the distribution over which the expectation is evaluated from $\rho_{exp}$ to the on-policy distribution for the learned policy $\rho_\pi$:
\begin{equation}
\label{eqn:dagger_loss}
\mathcal{L}_{DAG} = \mathbb{E}_{s \sim \rho_{dag}} \Big[ L(\pi_E(\cdot|s), \pi(\cdot|s))\Big],
\end{equation}
where $\rho_{dag}$ gradually shifts from $\rho_{exp}$ to $\rho_\pi$. In the limit, this approach learns a policy that imitates the expert along learned policy trajectories. 

\noindent
\textbf{AGGREVATE:} Whereas DAGGER clones the policy, AGGREVATE~\cite{Ross2014aggrevate} clones the value function using
\begin{equation}
\label{eqn:loss_aggrevate}
\mathcal{L}_{AGG} = \mathbb{E}_{s \sim \rho_{dag}} \Big[ L \big( Q(s,a), \hat{Q}(s,a) \big) \Big],
\end{equation}
where $\hat{Q}(s,a) = Q_E(s,a)$. Since AGGREVATE clones the value function, it is possible to use AGGREVATE as a preprocessing step to TD learning. The cloning algorithm proposed in this paper, $PQC$, can be viewed as a variant of AGGREVATE.

\noindent
\textbf{DQfD:} One problem with AGGREVATE is that it does a poor job reproducing the expert's policy. Equation~\ref{eqn:loss_aggrevate} minimizes loss with respect to $Q_E$, but it ignores the policy $\pi_E$. The result is that while $Q$ may be a good approximation of $Q_E$, the induced policy $\pi(s) = \argmax_{a \in A} Q(s,a)$ may be very different from the expert policy due to small errors in the approximation of $Q$. DQfD~\cite{hester2018dqfd} solves this problem by incorporating an additional large margin loss term that reduces the approximated value of non-expert actions
\begin{equation}
\label{eqn:largemargin}
    \mathcal{L}_{LM} = \max_{a \in A} \big[ Q(s,a) + l(a,\pi_E(s)) \big] - Q(s,\pi_E(s)),
\end{equation}
where $l(a,\pi_E(s))$ is a large constant positive number when $a \neq \pi_E(s)$ and zero otherwise. In state $s$, $\mathcal{L}_{LM}$ is large when the maximum action is different than the expert action. In order to reduce this error, the optimizer will decrease the values for non-expert $(s,a)$ pairs. Our version of DQfD combines $\mathcal{L}_{LM}$ with a standard TD loss: $\mathcal{L}_{dqfd} = \mathcal{L}_{TD} + \mathcal{L}_{LM}$.~\footnote{The loss proposed in~\cite{hester2018dqfd} actually contains two TD loss terms, the large margin loss above, and a regularization term, but we use this simplified version to make the algorithm more comparable to the others described in this section.} A key advantage of DQfD over AGGREVATE is that it learns an action value function for which the greedy policy is equal to the expert policy.

\noindent
\textbf{ADET:} Accelerated DQN via Expert Demonstrations (ADET)~\cite{adet} is another approach to the problem of small $Q$ estimation errors leading to a failure to reproduce the expert policy. However, rather than adding the large margin loss of Equation~\ref{eqn:largemargin}, ADET adds the DAGGER cross entropy loss on the policy instead: $\mathcal{L}_{ADET} = \mathcal{L}_{TD} + \mathcal{L}_{DAG}$, where the DAGGER loss is evaluated after applying a softmax to the Q-function output.

\section{Visual Servo Problem}

\noindent
\textbf{Problem Statement:} We assume we are given a discrete time system that includes a robotic hand or end-effector, a camera, and an environment that contains arbitrary objects placed randomly (see Figure~\ref{fig:intro}). Define $S$ to be a state space that spans all feasible end-effector poses and environment configurations. Let $A$ denote the action space that spans all feasible end-effector displacements on a single time step. Let $Z$ denote the observation space that spans the set of images that can be perceived by the camera. At the beginning of each episode, state (i.e. environmental configuration and end-effector pose) is randomly sampled from an initial distribution, $s \sim \rho_0$, and the agent must attempt to move an object grasped by the end-effector into a desired pose relative to the environment. The agent must find a policy $\pi : Z \mapsto A$ over the observation space that minimizes the expected time to reach a goal state for the given system. Importantly, we assume that the true state $s \in S$ is always hidden at test time. The agent only ever observes the camera images $z \in Z$.

\vspace{0.15cm}
\noindent
\textbf{Novel environment assumption:} A key assumption is that each experience with the world will take place with a novel configuration of ``clutter objects'' (Figure~\ref{fig:intro}). At the beginning of each episode, several objects uniformly randomly sampled from a database of clutter objects are placed in random positions and orientations in the scene (Figure~\ref{fig:experimental_scenarios}). The task objective remains the same -- to move the grasped object into a desired pose with respect to the environment -- but the presence of clutter makes the problems of visual perception and obstacle avoidance more challenging.

\vspace{0.15cm}
\noindent
\textbf{Train/test information asymmetry assumption:} We assume that the agent has access to a fully modeled MDP and a simulator during training. Specifically, we assume that the agent has access to the transition model $T : S \times A \mapsto S$ and the observation model $h : S \mapsto Z$ and that the agent observes the full Markov state $s \in S$ at each time step. During training, the simulator produces state and simulated camera observations. At test time, we assume that the agent does \emph{not} observe the full state. Instead, the agent only has access to camera observations $z \in Z$.

This paper focuses on visual servo problems that do not involve contact. Also, we largely ignore the domain transfer problem, relying on an off-the-shelf method~\cite{isola2017image} to bridge the sim2real gap.

\section{Penalized Q Cloning (PQC)}

\begin{wrapfigure}{r}{0.25\textwidth}
  \begin{center}
    \includegraphics[width=0.25\textwidth]{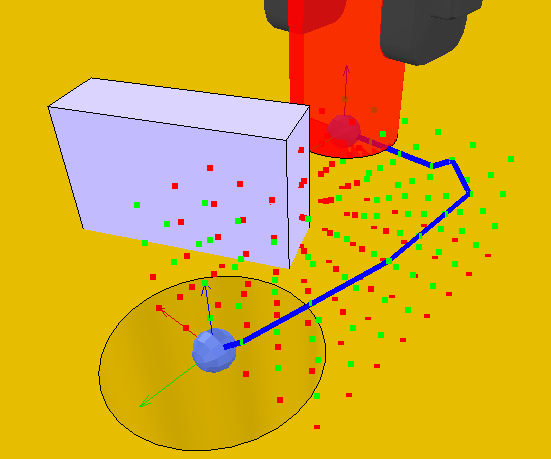}
  \end{center}
  \caption{Illustration of full state motion planning scenario.}
  \label{fig:full_state_motion_plan}
\end{wrapfigure}

In view of the train/test information asymmetry assumption (above), notice that there are really two relevant MDPs: 1) a fully modeled MDP over the underlying state space $S$ and action space $A$; and 2) an unmodeled MDP over the observation space $Z$ and action space $A$. We want to find a policy for the unmodeled MDP, but we can only plan in the modeled MDP. Our approach is therefore to use a full state motion planner to generate an ``expert'' policy and value function for the fully modeled MDP and then to project those solutions onto the unmodeled MDP using behavior cloning.

\subsection{Full state motion planner as the expert}

First, we need to solve the fully modeled MDP for an expert value function and policy. In general, this solution is found using some form of approximate dynamic programming. However, in our particular case, since the modeled MDP corresponds to a collision free motion planning problem, we can use any standard approximately optimal motion planner, e.g. sPRM~\cite{kavraki1996probabilistic}, RRT$^*$~\cite{karaman2011sampling}, etc. In our experiments, we use Djikstra's algorithm over a regular 3D grid of end-effector positions. Edges in the graph correspond to Euclidean distances between end-effector positions (in meters) plus a penalty for approaching obstacles. This penalty is zero for distances greater than 2 cm and linearly increases to a cost of 1 as distance drops to zero. Shortest paths to goals in the graph correspond to optimal solutions to the modeled MDP, assuming a reward function that applies negative rewards equal to the costs described above. Solutions found by the planner provide two pieces of information to the behavior cloner: an approximately optimal policy $\pi_E$ and an approximately optimal value function $Q_E$. Figure~\ref{fig:full_state_motion_plan} shows an example of a trajectory found using this method.

\subsection{Cloning a full state motion planner}
\label{sect:approach_cloning}

Now that we have found an approximately optimal policy and value function for the modeled MDP, we need to project these solutions onto the unmodeled MDP using behavior cloning. We do the following:

\noindent
\textbf{(a) Sample trajectories from expert:} First, we sample trajectories from the expert by sampling an initial state $s_0 \sim \rho_0$ and then rolling out the expert $\pi_E$ until termination of the episode. For each action, we simulate the resulting next state and query the expert for an approximate q-value, $Q_E(s,a)$. This results in a sequence of experiences $(s_1,a_1,q_1), \dots, (s_n,a_n,q_n)$, where $q_i = Q_E(s_i,a_i)$, that are accumulated in a dataset $\mathbb{D}$.

\noindent
\textbf{(b) Augment dataset with supervision for all actions:} Because we have a small number of actions and the expert is implemented by a planner, we can generate supervision for all feasible actions from a given state, not just those executed by the simulator. We do this for all states in $\mathbb{D}$. Specifically, for each experience $(s,a,q) \in \mathbb{D}$, we generate $|A|-1$ additional experiences $\{(s,a',Q_E(s,a')) | a' \in A \setminus a\}$ and add these experiences to $\mathbb{D}$.

\noindent
\textbf{(c) Apply a penalty to non-expert actions:} Unlike an approach like DAGGER~\cite{ross2011dagger} which clones the policy directly, here we are cloning the value function. This exposes us to a key failure mode: we may learn a good estimate of $Q_E$ while still estimating $\pi_E$ poorly. Under ideal conditions, our estimate $Q$ is exactly equal to $Q_E$: $Q(s,a) = Q_E(s,a), \; \forall s,a \in S \times A$. In this case, the greedy policy of the learned value function is equal to expert policy: $\argmax_a Q(s,a) = \argmax_a Q_E(s,a) = \pi_E(s)$. However, since we are using a deep neural network to approximate $Q_E$, we can expect small errors. This is a problem because even small errors can result in a substantial divergence between $\argmax_a Q(s,a)$ and $\pi_E(s)$. To combat this, we set the action values of non-expert actions to a fixed value $c$ where $c < \min_{s,a \in S \times A} Q_E(s,a)$ (line 10 of Algorithm~\ref{alg:pqc}).

\begin{algorithm}
\caption{Batch Penalized Q Cloning (Batch PQC)}\label{alg:pqc}
\begin{algorithmic}[1]
\State $\mathbb{D} \gets \varnothing$
\While{more episodes to execute}
    \State $s_0 \sim \rho_0$
    \For{$t \in [0,T-1]$} \Comment{iterate over time steps}
        \State $a_t = \pi_E(s_t)$
        \For{$\forall a \in A$}
            \If{$a = a_t$}
                \State $q \leftarrow Q_E(s_t,a)$
            \Else
                \State $q \leftarrow c$ \Comment{apply penalty}
            \EndIf                    
            \State $\mathbb{D} \gets \mathbb{D} \cup \{(s,a,q)\}$
        \EndFor 
        \State $s_{t+1} \leftarrow T(s_t,a_t)$
    \EndFor
\EndWhile
\State Find $Q$ that minimizes $\mathcal{L}_{PQC}$
\State \textbf{Return} $Q$
\end{algorithmic}
\end{algorithm}

After generating the dataset $\mathbb{D}$ using the above, we use standard SGD-based methods to optimize
\begin{equation}
\mathcal{L}_{PQC} = \mathbb{E}_{(s,a,q) \sim \mathbb{D}} \Big[ L(Q(s,a), q) \Big],
\end{equation}
where $L(\cdot)$ is an appropriate loss. We call this method \emph{batch penalized Q cloning} or batch PQC for short. The full algorithm is shown in Algorithm~\ref{alg:pqc}.

\noindent
\textbf{Relationship of this cloning method to prior work:} This approach to behavior cloning draws elements from at least two different pieces of prior work. First, since we are cloning the value function rather than the policy, our method can be viewed as a form of AGGREVATE~\cite{Ross2014aggrevate}. Second, the fact that we administer a penalty to non-expert actions is similar to what is done in DQfD~\cite{hester2018dqfd} and ADET~\cite{adet}. However, since both of those methods use TD learning, they must add an additional term into the loss function in order to achieve this. DQfD adds the relatively complex large margin loss term. ADET adds a cross entropy term between the policy implied by the value function and the expert policy. In contrast, since our method uses a fully supervised target, we can simply reduce the supervised target $q$ value without the additional loss term. We experimentally compare our approach to AGGREVATE and DQfD in Section~\ref{sect:experiments}. 

\noindent
\textbf{Algorithm variations:} We also consider a couple of dimensions of variation on the basic algorithm described above.

\begin{itemize}
    \item \underline{Online PQC:} Instead of training over a large batch of expert rollouts, \emph{online} PQC uses a DAGGER-like rollout schedule, i.e. it alternates between rollouts from the expert policy and rollouts from the learned policy according to a schedule. Early in training, most rollouts come from the expert. Later, most rollouts come from the learned policy. Our experiments indicate that this method outperforms on the training set but underperforms in novel environments due to overfitting.
    
    \item \underline{Relative penalty PQC:} In this variant, we replace the constant penalty (line 10 of Algorithm~\ref{alg:pqc}) with a relative penalty: $q \leftarrow Q_E(S_t,a) - l$ where $l$ is the margin by which to penalize the non-expert action. This version of the algorithm is more simliar to DQfD~\cite{hester2018dqfd}. Our experiments indicate that it outperforms on the training set but underperforms in novel environments due to overfitting.
    
\end{itemize}

\subsection{Caching visual observations}
\label{sect:caching_method}

One of the challenges of training in simulation is that it can be computationally expensive to simulate depth images of simulated worlds. In principle, each transition added to the dataset $\mathbb{D}$ requires simulating an image of a novel scene from a novel perspective. In order to speed this up this process, we precompute observations for a large set of clutter configurations. Specifically, we precompute the image that would be observed from each node in the motion planning graph. Then, when rolling out a trajectory during training, we simply recall the observation that corresponds to a given end-effector pose rather than recomputing it. This enables us to roll out as many trajectories as we want from a given scene. As long as this graph is not too large, this approach works well. Using this procedure, since we no longer create a new scene each time a new episode executes, we must select some number of scenes in advance for which to generate data. In our experiments, we simulate observations for at least 500 different scenes where each scene contains randomly selected clutter in a random task setting. In the experiments, we evaluate the effect of increasing the number of scenes used to create this training set (Figure~\ref{fig:perf_scenes}).

\subsection{Finetuning using TD learning}

After learning a policy using the behavior cloning methods described above, we follow up with additional training using pure TD learning (DQN in our case, but other algorithms should accomplish the same thing). We view this as a finetuning step: the cloning phase (pretraining) learns an approximately correct value function and the TD learning phase (finetuning) makes small adjustments to improve performance. TD learning has the potential to improve upon a suboptimal planner as well as correct value function inaccuracies caused by the penalty applied to non-expert actions. Our experiments indicate that this phase of training generally helps, improving performance of policies produced by most cloning algorithm variations over both the test set and training set.

\section{Experiments in simulation}
\label{sect:experiments}

\subsection{Experimental setup}

\begin{figure}
      \centering
      \subfigure[]{\includegraphics[width=1.5in]{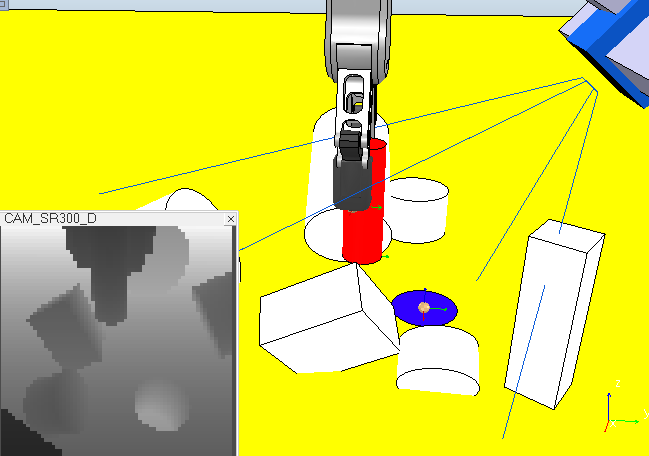}}
      \subfigure[]{\includegraphics[width=1.5in]{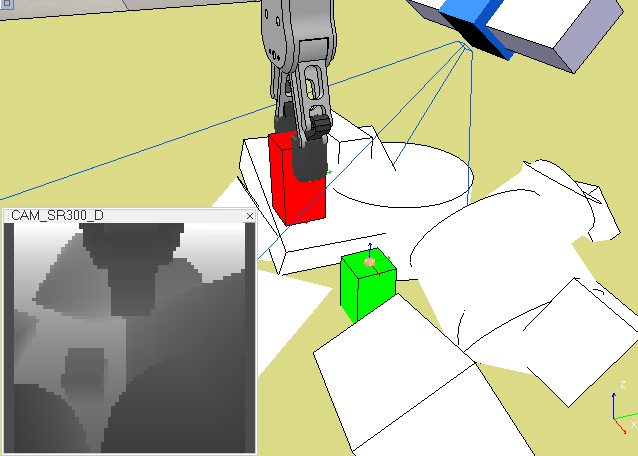}}
      \caption{(a) peg insertion scenario. (b) block stacking scenario.}
      \label{fig:experimental_scenarios}
      \vspace{-0.4cm}
\end{figure}

We evaluate PQC against several algorithm variations and baselines on two visual servoing tasks: a peg insertion task and a block stacking task (Figures~\ref{fig:experimental_scenarios}a and b, respectively). In peg insertion, the robot starts execution with the peg in its hand and must move it until it reaches a goal pose just above the hole. The block stacking task (Figure~\ref{fig:experimental_scenarios} b) is similar except that the robot starts with a block in its hand and must move it to a goal pose just above a second block. To solve these tasks, the agent must learn to servo while avoiding obstacles. In addition, the block stacking task requires the policy to determine \emph{which} block to stack upon by matching the visual image of the grasped block with the blocks that are visible in the scene.

In both tasks, the scene is populated by randomly placed clutter that serves both as visual complexity and as obstacles that must be avoided during servoing. In each scene of peg insertion, both peg and hole size are sampled uniformly randomly (peg smaller than hole) and the peg is displaced in the hand with a small random offset. Similarly, in each scene of block stacking, both block sizes are sampled uniformly randomly, the grasped block is offset by a random displacement, and a small uniform random rotation is applied to the block to be stacked upon. Figure~\ref{fig:experimental_scenarios} illustrates the depth images observed by the agent. Note that they include both the grasped object and the scene. The expert is implemented by a planner that uses Djikstra's algorithm over a $21 \times 21 \times 12$ grid of positions (fixed orientation) with vertices at 1cm intervals (a total of $5k$ vertices, see Figure~\ref{fig:full_state_motion_plan}).

\begin{figure*}[t]
      \centering
      \subfigure[Peg-insertion; training set]{\includegraphics[width=1.75in]{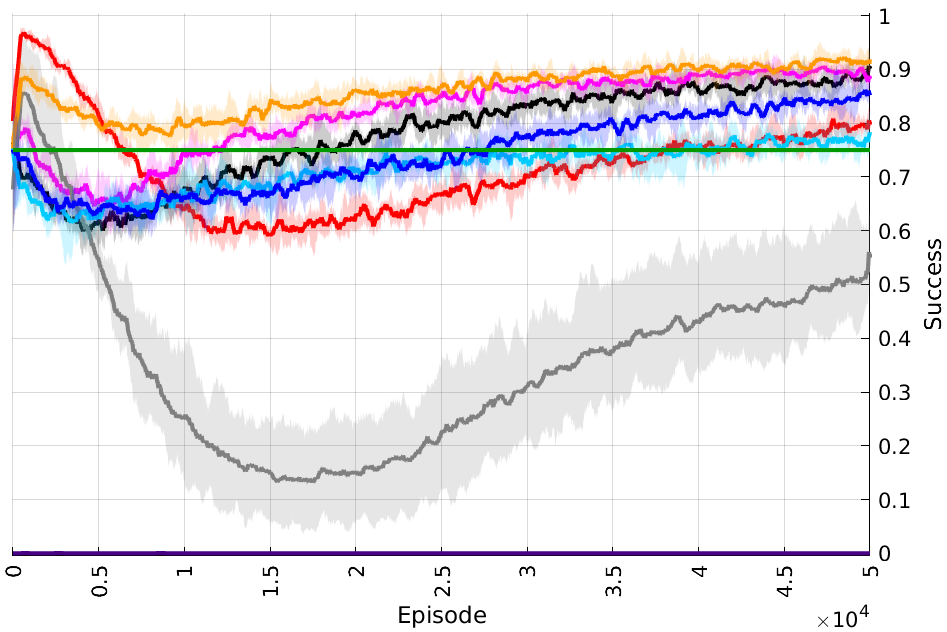}}
      \subfigure[Peg-insertion; holdout test set]{\includegraphics[width=1.75in]{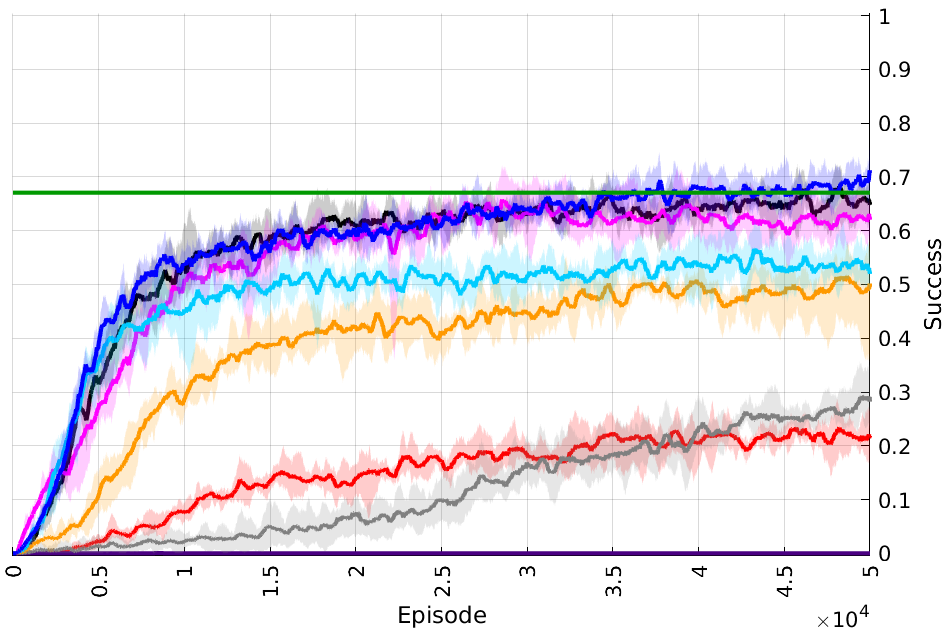}}
      \subfigure[Block-stacking; training set]{\includegraphics[width=1.75in]{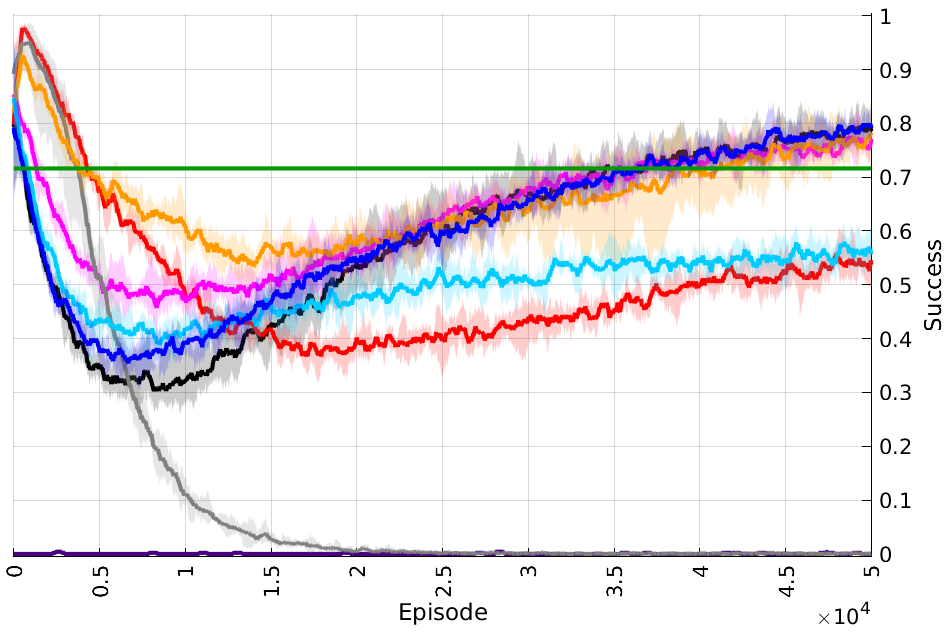}}
      \subfigure[Block-stacking; holdout test set]{\includegraphics[width=1.75in]{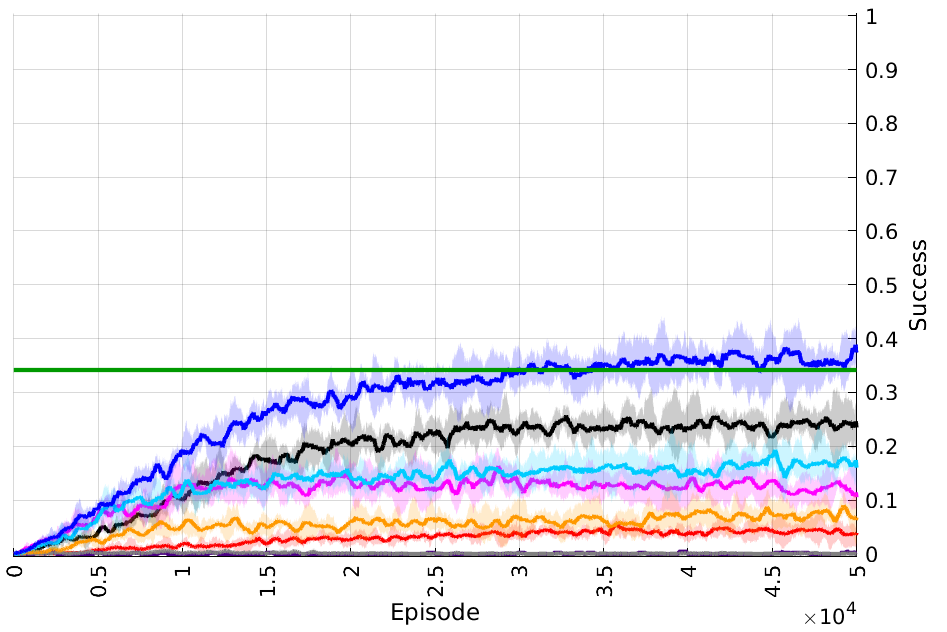}}
      \caption{Success rate as a function of training episode during cloning. (a,b) are peg-insertion. (c,d) are block-stacking. (a,c) show success rates on the training set. (b,d) show success rates on the holdout test set. Colors explained in Section~\ref{sect:cloning_results}.}
      \label{fig:clone_results}
      \vspace{-0.2cm}
\end{figure*}

\subsection{Training details}
\label{sect:training_details}

For each task, we created a dataset with 50k episodes by generating 500 scenes and rolling out 100 episodes per scene. Each scene was populated by random clutter and random peg/hole/block sizes and positions as described above. Since each episode is approximately 10 to 15 steps long, this dataset contains at least 500k transitions. For testing purposes, we created a second holdout dataset with 50k episodes created by generating 100 additional different scenes. We evaluated each algorithm on both the training set and the holdout set.

We compare batch PQC with several online algorithms. The online algorithms train using data produced using a policy with a DAGGER-like schedule. Initially, the online policy rolls out 80\% of its transitions from the expert policy and 20\% from a uniform random policy. The proportion of expert and random transitions decays exponentially (but with the same 80\%/20\% proportion) until all transitions are sampled from the learned on-policy distribution after 25k episodes. Batch PQC trains for eight epochs on the 50k episodes rolled out from the expert (i.e. the planner). The online algorithms train for 50k episodes of data generated by rolling out the online policy as described above. All batch sizes are 64.

\subsection{Comparisons with ablations and baselines}
\label{sect:cloning_results}

Figure~\ref{fig:clone_results} compares the performance of fixed penalty batch PQC with a variety of ablations and baselines on our two tasks, peg-insertion and block-stacking. A task is considered to have succeeded if the agent reaches a position within 1cm of the goal.

\vspace{0.1cm}
\noindent
\underline{1. Batch PQC (green)}: Version of Batch PQC shown in Algorithm~\ref{alg:pqc}. In this experiment we set $c=-0.5$ (line 10). Batch training was for eight epochs on the 50k episode training set (see Section~\ref{sect:training_details}).

\vspace{0.1cm}
\noindent
\underline{2. Online PQC (blue)}: Same as batch PQC above except that it is trained online using a DAGGER-like rollout schedule. Trained for 50k episodes.

\vspace{0.1cm}
\noindent
\underline{3. Online PQC with no penalty (red)}: Ablation. Same as online PQC above except that there is no penalty for non-expert actions.

\vspace{0.1cm}
\noindent
\underline{4. Online PQC one action update (cyan)}: Ablation. Same as online PQC except that we delete lines 9 and 10 of Algorithm~\ref{alg:pqc} (only update the value of the action selected for execution). 

\vspace{0.1cm}
\noindent
\underline{5. Online PQC, relative penalty (orange):} Same as online PQC except that we change line 10 of Algorithm~\ref{alg:pqc} to: $q \leftarrow Q_E(s_t,a) - l$, where $l = 0.2$ (the same margin used by DQfD).

\vspace{0.1cm}
\noindent
\underline{6. DAGGER (black):} Baseline. Classic DAGGER algorithm implemented using the standard cross entropy loss.

\vspace{0.1cm}
\noindent
\underline{7. DQfD with DAGGER schedule (magenta):} Baseline. This version of DQfD uses a single TD loss term plus the large margin loss (margin equal to $0.2$ in this experiment) weighted at one tenth the contribution of the TD term (this combination gave us the best performance). It was trained using the DAGGER schedule.

\vspace{0.1cm}
\noindent
\underline{8. DQN on-policy (purple)}: Baseline. DQN trained entirely on-policy.

\vspace{0.1cm}
\noindent
\underline{9. DQN with DAGGER schedule (grey)}: Baseline. DQN trained off-policy using the DAGGER schedule.

For more details on methods \#1 -- \#5, see Section~\ref{sect:approach_cloning}. For more details on methods \#6 -- \#9, see Section~\ref{sect:background}

Based on these results, a few things are immediately clear. First, on-policy DQN (purple) underperforms significantly. This justifies our fundamental choice to clone an expert rather than use model-free learning. DQN performance can be improved by providing expert demonstrations, i.e. DQN with the DAGGER schedule (grey), but it still underperforms other methods. Second, our proposed methods, batch PQC (green) and online PQC (blue) both perform similarly or slightly worse than two baselines, DQfD (magenta) and DAGGER (black), on the training set. However, they outperform these two baselines on the holdout test set. This is particularly true for the block stacking task which is harder than peg insertion because it requires observing the size of the grasped block in order to determine where to stack. This suggests that PQC generalizes better to new scenes with different clutter configurations. Third, the results indicate that the two ablations, online PQC with no penalty (red) and online PQC with one action update (the two ablations of online PQC), underperform, suggesting that both these elements of the algorithm are important. Finally, if we compare online (fixed penalty) PQC versus relative penalty online PQC, we see that the relative penalty version outperforms on the training set but significantly underperforms on the test set. This suggests that the fixed penalty version generalizes better to novel scenes.

\subsection{Effect of adding more training scenes}

\begin{figure}
  \begin{center}
    \includegraphics[width=0.4\textwidth]{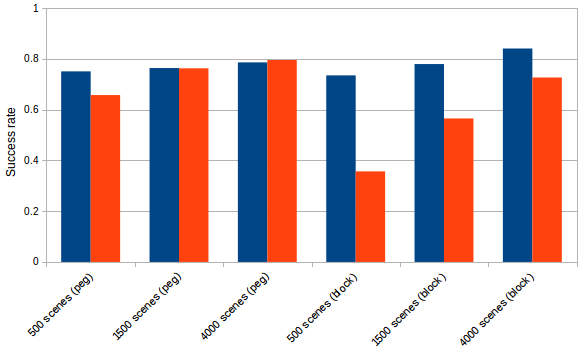}
  \end{center}
  \vspace{-0.2in}
  \caption{Success rate achieved by fixed penalty batch PQC when training with different numbers of scenes. Blue: performance on training set. Red: performance on holdout test set. The left three pairs of bars are for peg insertion. The right three are for block stacking.}
  \label{fig:perf_scenes}
\vspace{-0.5cm}
\end{figure}

Recall from Section~\ref{sect:caching_method} that we performed online rollouts using cached visual observations. This accelerated online learning because it reduced the number of (expensive) visual rendering operations performed by the simulator during training. However, comparing the results of Figure~\ref{fig:clone_results}a and c with those in b and d suggests that there is a significant performance drop when evaluating on holdout test data rather than training data, especially on the blocks task. This suggests that 500 different scenes are not enough to generalize to scenes with new clutter configurations. To test this, we repeated the fixed penalty batch PQC trial with training sets generated with 1500 and 4000 novel scenes instead of just 500. Figure~\ref{fig:perf_scenes} shows the result. Performance on the training set improved marginally but performance on the holdout test set improved dramatically, especially for the block stacking task. Why does cloning seem to require so many novel scenes? If the only effect of clutter was to create scenes that were more complex visually, then 500 should be sufficient. However, our tasks require learning a policy that can avoid colliding with the clutter objects. The blocks task is particularly challenging because the agent must match the image of the grasped block with a block in the scene. These tasks appear to require training data generated from a large number of unique scenes.

\begin{table}
\begin{center}
\begin{tabular}{ |c|c|c| } 
 \hline
 Method & Peg/Hole & Blocks \\ 
 \hline
Fixed penalty batch PQC & 649s (data gen) & 519s (data gen) \\ 
& +104s (train) & + 160s (train) \\ 
\hline
Fixed penalty online PQC & 5009s & 4625s \\ 
 \hline
\end{tabular}
\end{center}
\caption{Average training times for a 50k episode dataset: batch versus online training.}
\vspace{-0.65cm}
\label{table:timing}
\end{table}

The above discussion is critical to understanding why the batch version of PQC is superior to the online version in our experimental setting. In our experiments, the fixed penalty batch PQC trains several times faster per optimization step than the online version (see Table~\ref{table:timing}, results averaged over 5 runs). This is because batch training does not require running the simulator periodically during optimization -- it just samples from a dataset. Without using the batch version, we \emph{cannot} reasonably train with the large datasets needed to generalize well to scenes with novel clutter.

\subsection{Prioritized Experience Replay}

\begin{wrapfigure}{r}{0.25\textwidth}
\vspace{-0.5cm}
  \begin{center}
    \includegraphics[width=0.25\textwidth]{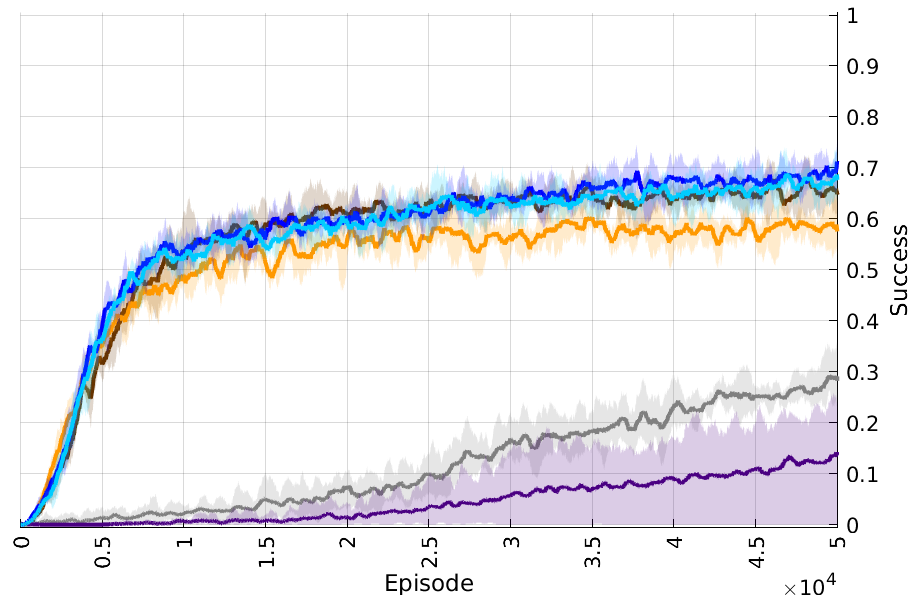}
  \end{center}
  \caption{DAGGER and online PQC with and without PER. Blue and cyan show PQC performance with and without PER (resp.) Black and grey similarly show DAGGER performance.}
  \label{fig:per}
\end{wrapfigure}

Prioritized experience replay (PER) is an approach to biasing minibatch sampling in DQN~\cite{schaul2015prioritized}. Instead of sampling batches at random from the replay buffer, PER samples them with a probability proportional to a power of the TD error and corrects sample bias using importance sampling. We adapted this idea to the imitation learning setting by prioritizing experiences using the loss for whatever method was being used. For example, in DAGGER, we prioritize by magnitude of the cross entropy loss for a particular experience (the other elements of PER are the same). Figure~\ref{fig:per} shows a comparison between DAGGER and batch PQC, with and without PER. We conclude that PER can sometimes improve cloning performance and we have never seen it hurt performance in our scenarios. Therefore, all experiments in this paper use PER.

\subsection{Finetuning with TD Learning}

\begin{figure}
  \begin{center}
    \includegraphics[width=0.4\textwidth]{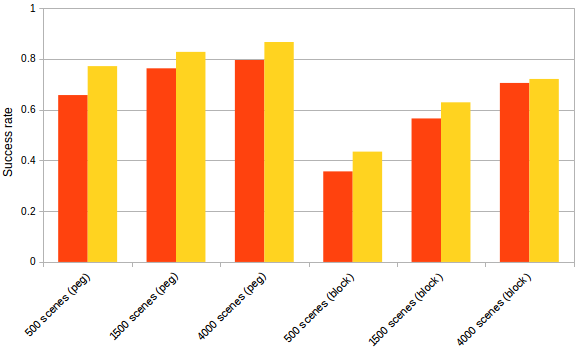}
  \end{center}
  \vspace{-0.2in}
  \caption{Success rates fixed penalty batch PQC agents on holdout test set before (red) and after (yellow) finetuning using DQN.  The left three pairs of bars are for peg insertion. The right three are for block stacking.}
  \label{fig:td_scenes}
\vspace{-0.5cm}
\end{figure}

After cloning, we ``finetune'' the policy using TD learning (standard DQN) on a new 500 scene dataset. The dataset used for TD learning is distinct from both the training set used for cloning and the holdout test set. TD learning follows the same DAGGER-like expert/on-policy schedule that was used for DQN earlier (see Section~\ref{sect:training_details}). Results are shown in Figure~\ref{fig:td_scenes}. They show that DQN results in a small improvement in performance on the holdout test set for both tasks.

Why does TD learning help? First, the expert policy that is cloned during pretraining is only approximately optimal -- so there is room for TD learning to improve upon it. Second, recall that the penalty term used in many of the cloning methods causes the agent to learn an incorrect value function, albeit one that supports a near-optimal policy. TD learning probably does a better job finding values that are nearly correct \emph{and} that support a near-optimal policy.

\section{Validation on a physical robot}

\begin{figure*}
  \centering
  \includegraphics[height=0.75in]{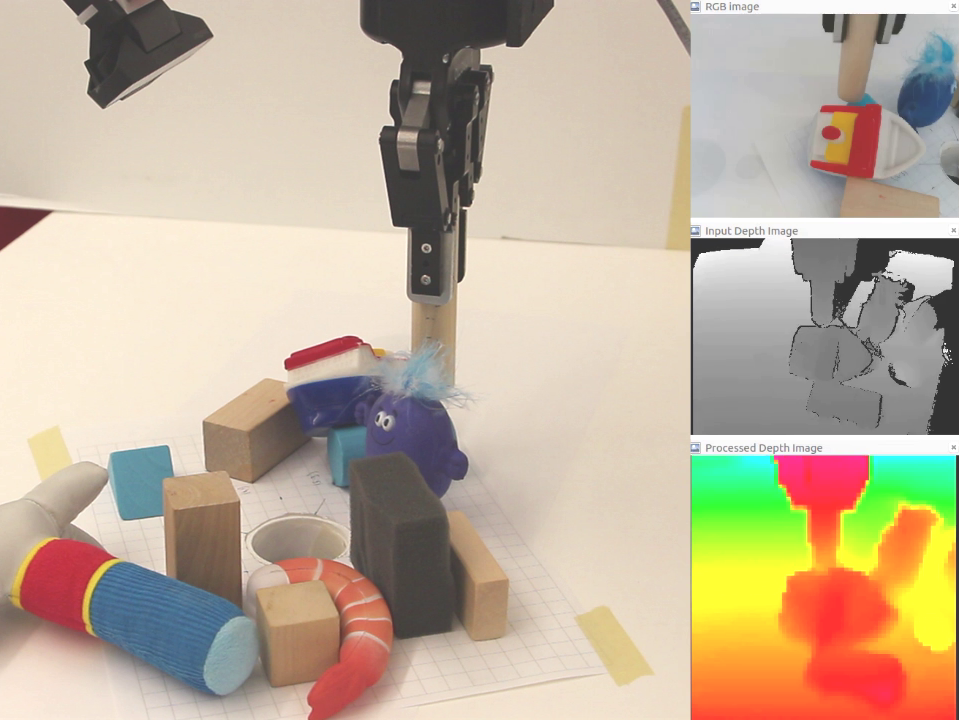}
  \hspace{0.25cm}
  \includegraphics[height=0.75in]{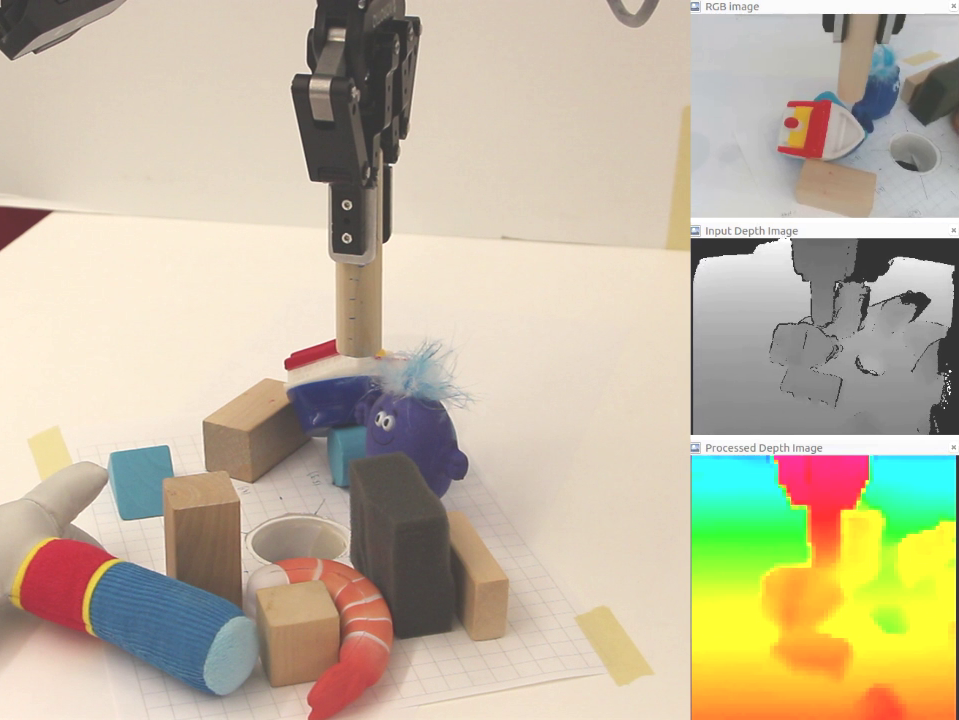}
  \hspace{0.25cm}
  \includegraphics[height=0.75in]{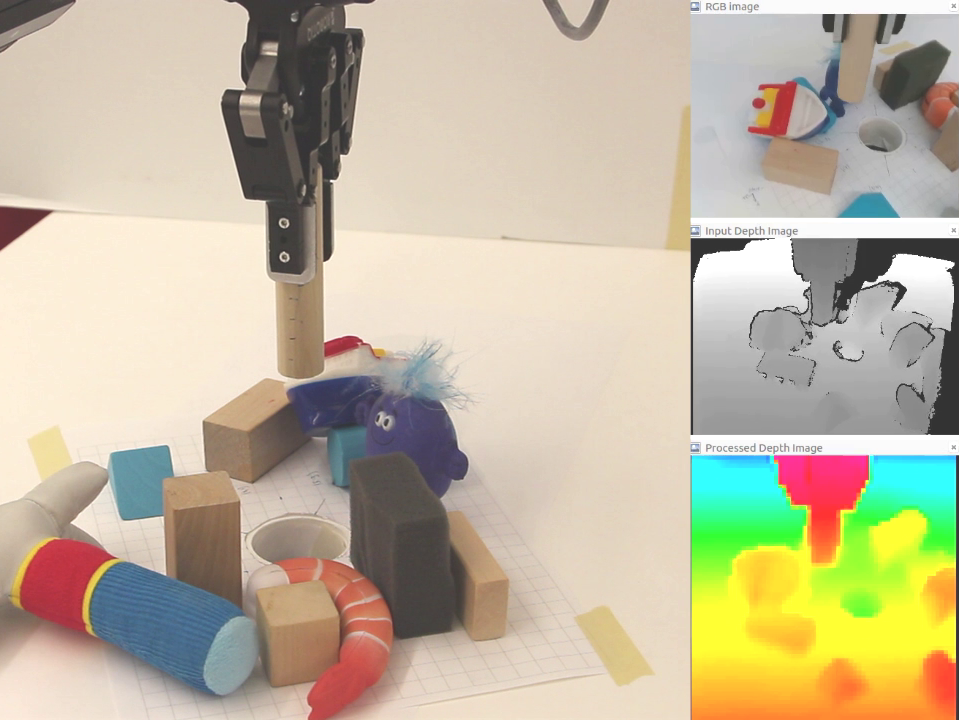}
  \hspace{0.25cm}
  \includegraphics[height=0.75in]{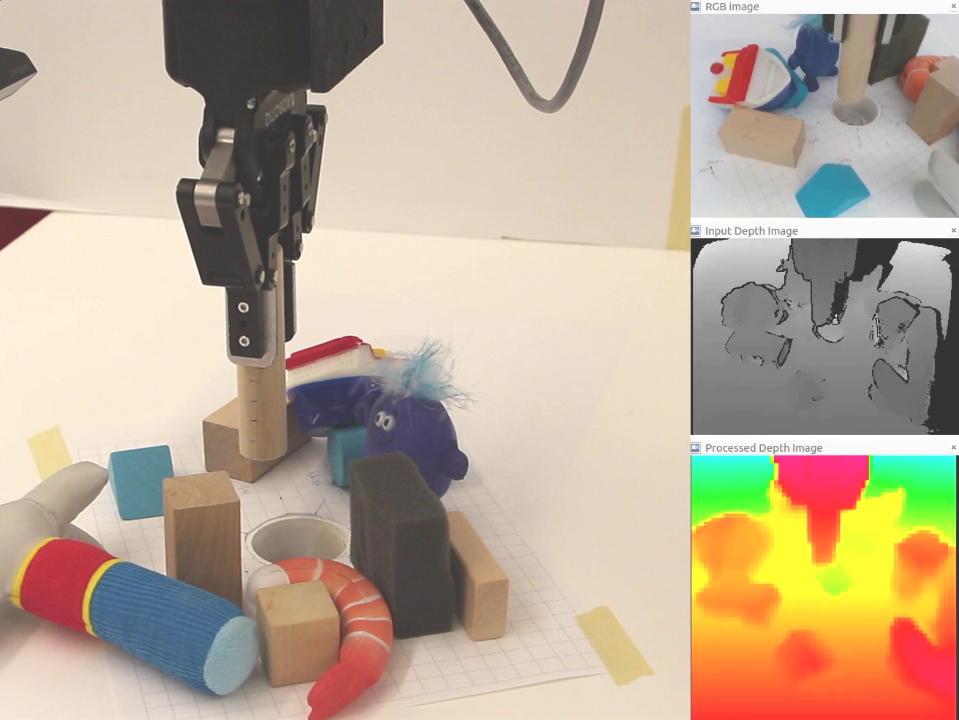}
  \hspace{0.25cm}
  \includegraphics[height=0.75in]{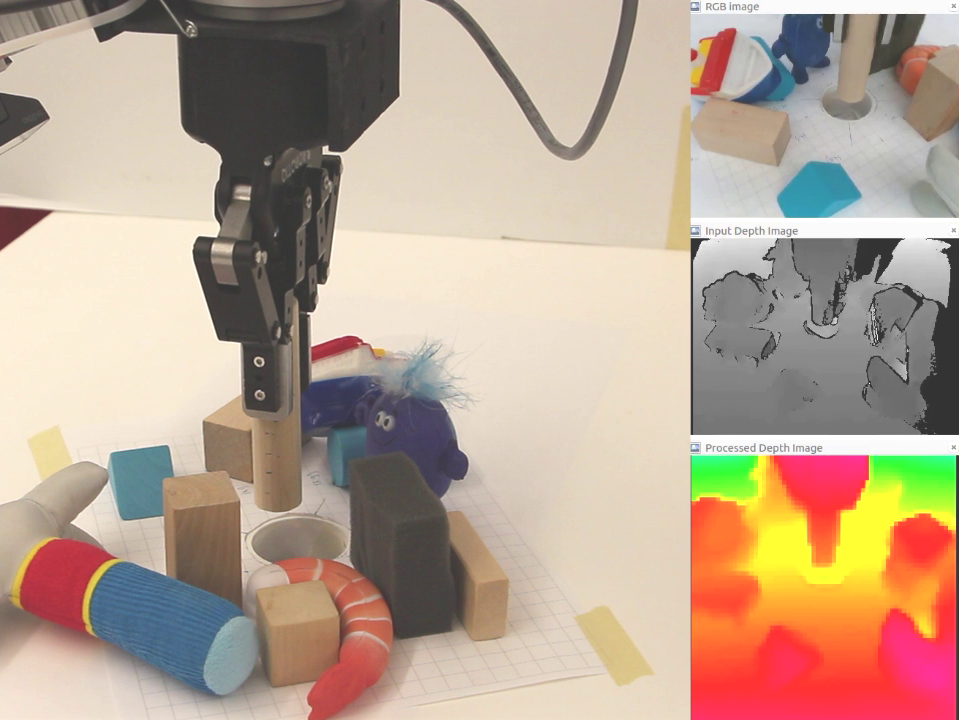}
  \hspace{0.25cm}
  \includegraphics[height=0.75in]{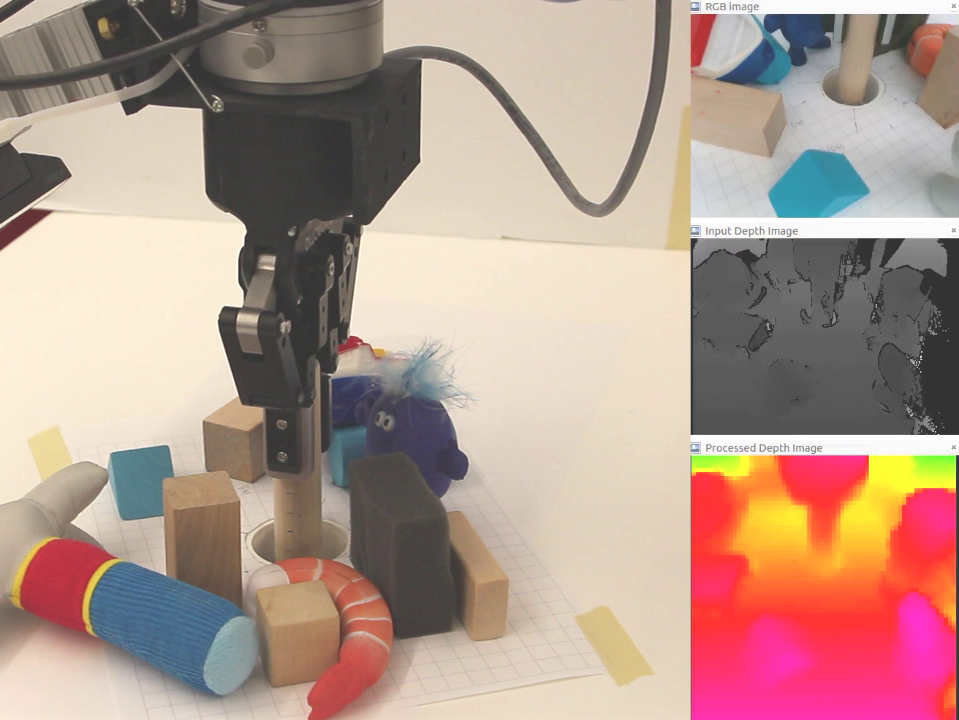}      
  \caption{Example of a single trial on our robot.}
  \label{fig:physical_trial}
\end{figure*}

\begin{figure}
      \centering
      \subfigure[Scene 1]{\includegraphics[width=1.6in]{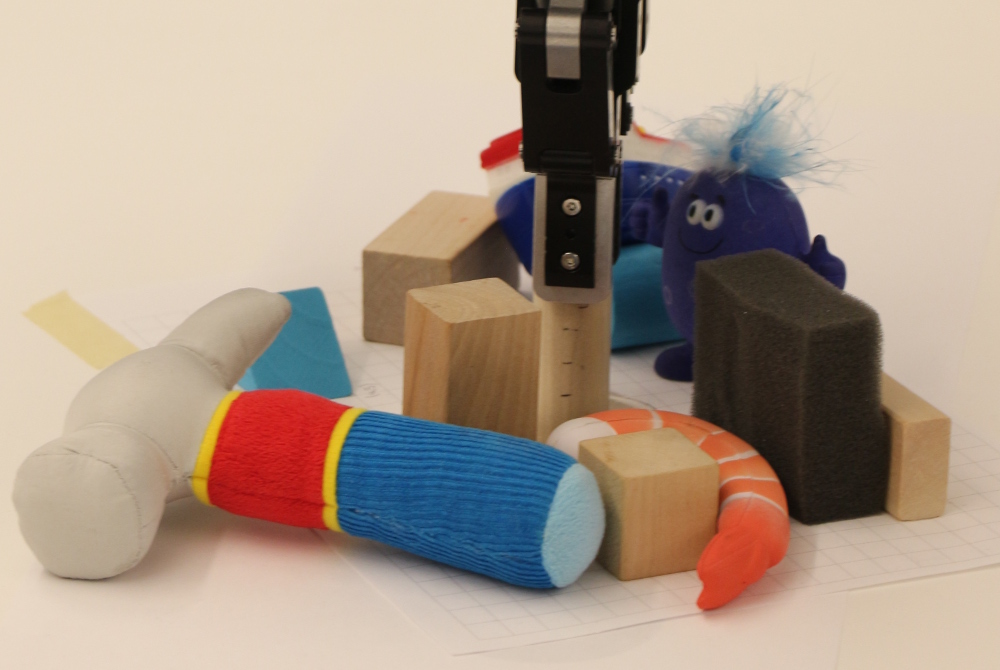}}         
      \subfigure[Scene 2]{\includegraphics[width=1.6in]{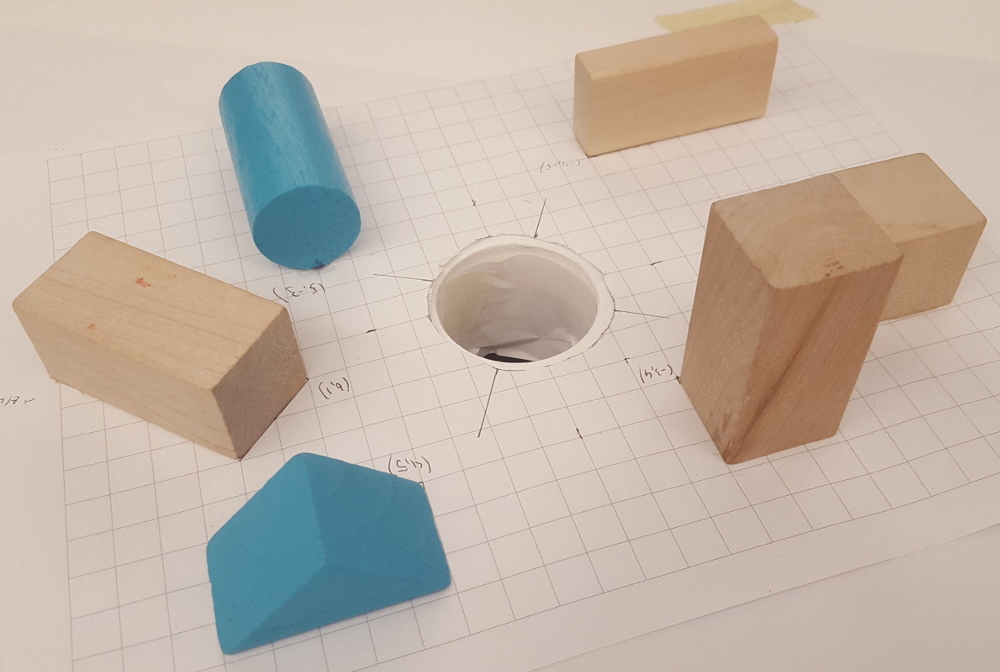}}\\      
      \subfigure[Scene 3]{\includegraphics[width=1.6in]{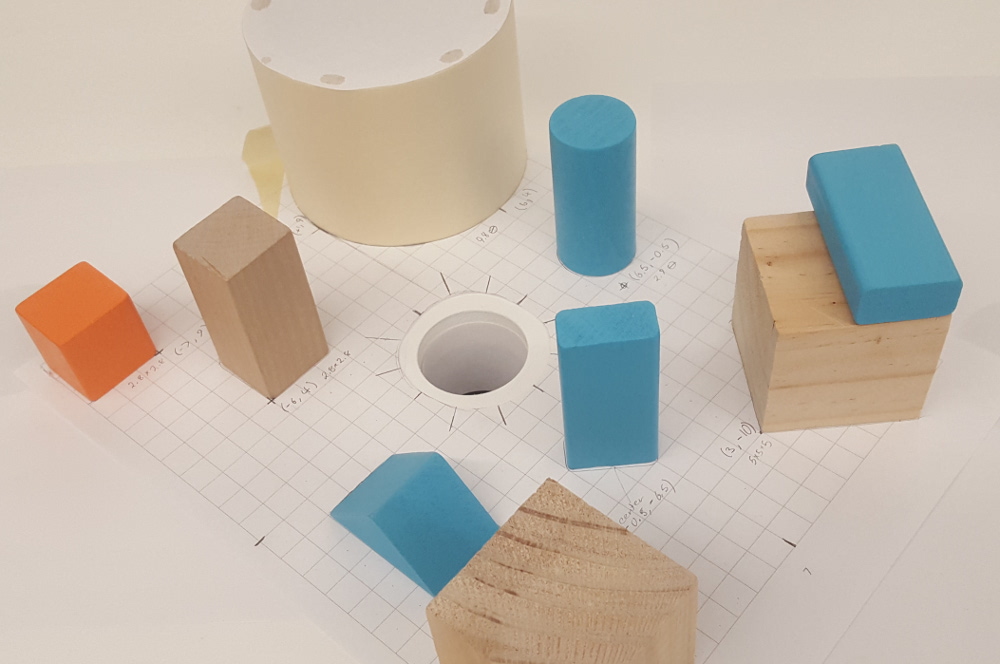}}
      \subfigure[Scene 4]{\includegraphics[width=1.6in]{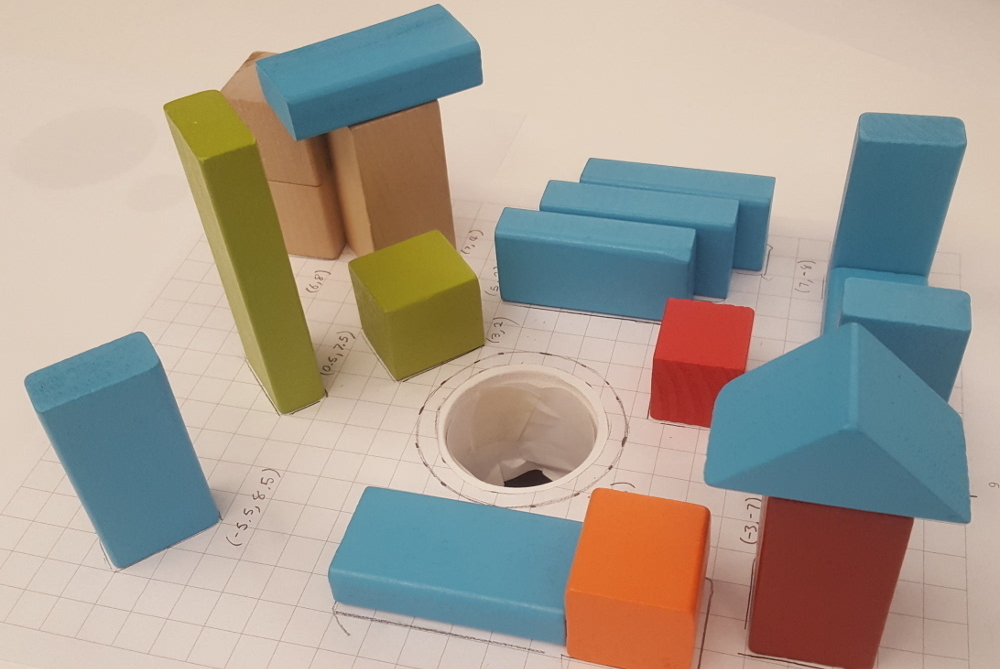}}
      \subfigure[Scene 5]{\includegraphics[width=1.6in]{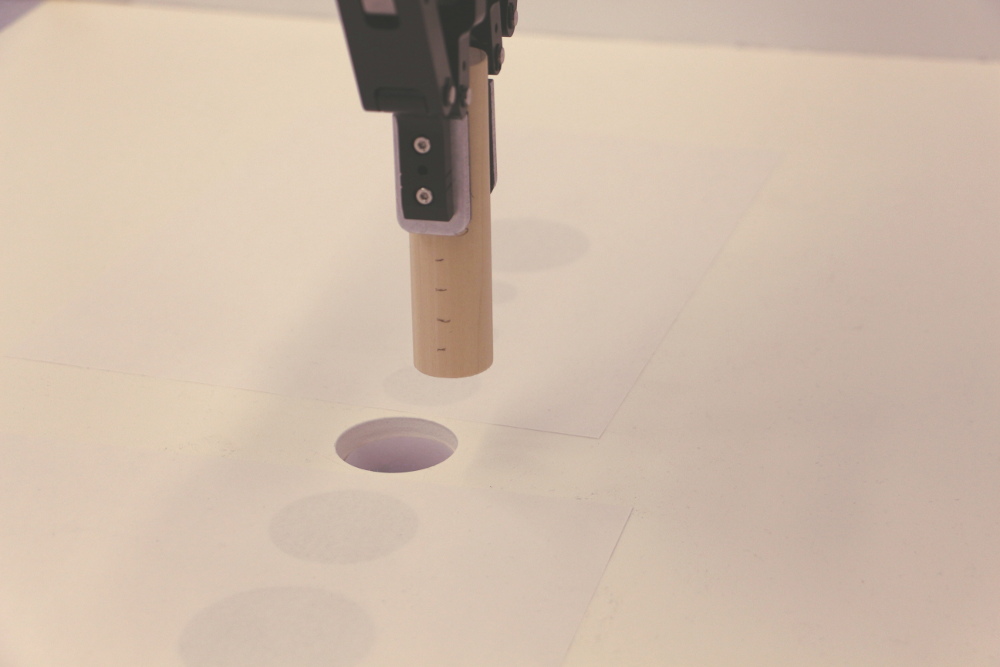}}      
      \caption{Different scenes for robot trials. Diameters of peg/hole: (a,b) 2.2cm/4.2cm, (c) 2.2cm/3.9cm, (d) 1.9cm/4.2cm, Scene 5 (no clutter): 2.54cm/3.9cm }
      \label{fig:exp_scenario}
\vspace{-0.5cm}
\end{figure}

We performed 100 proof-of-concept trials on the robotic system shown in Figure~\ref{fig:exp_scenario} for the peg-insertion task on five novel scenes containing novel objects placed arbitrarily. Because of the small number of trials, these experiments cannot measure performance precisely. However, they do demonstrate that the results from our evaluation in simulation (Figures~\ref{fig:clone_results},~\ref{fig:td_scenes}) roughly correspond to expected performance on a physical system. The experiments were performed using a Robotiq Two Finger gripper mounted on a UR5 robotic arm in a tabletop setting. Depth images were produced by an Intel SR300 depth sensor mounted near the robotic hand as illustrated in Figure~\ref{fig:intro}.

\begin{figure}
      \centering
      \subfigure[]{\includegraphics[width=1.1in]{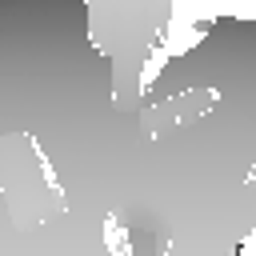}}
      \subfigure[]{\includegraphics[width=1.1in]{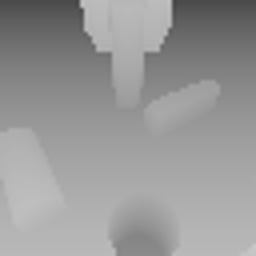}}
      \subfigure[]{\includegraphics[width=1.1in]{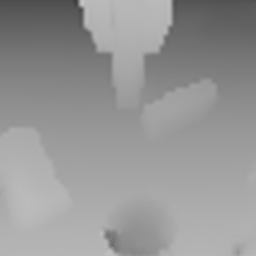}}
      \subfigure[]{\includegraphics[width=1.1in]{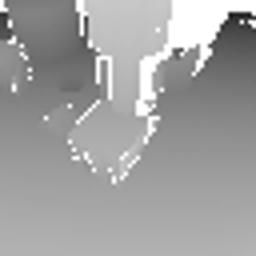}}
      \subfigure[]{\includegraphics[width=1.1in]{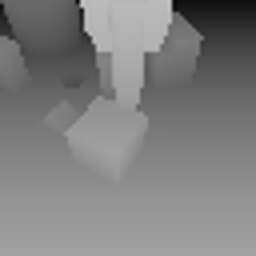}}
      \subfigure[]{\includegraphics[width=1.1in]{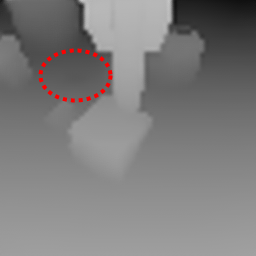}}      
      \caption{Examples of corresponding real and simulated depth images of gripper holding a peg. (a,d) real image; (b,e) simulated image; (c,f) prediction made by the pix2pix GAN of what the simulated image would look like given only the real image. (f) shows that it might fail to transform the hole (see area marked with red circle) because there is no objective to focus on it.}
      \label{fig:domain_shift}
\end{figure}

\subsection{Depth image domain shift}

A key challenge that is deemphasized in this work is the domain shift problem for depth images. The problem is that the observation model used by the simulator is not perfectly accurate, as illustrated in Figure~\ref{fig:domain_shift}. Figure~\ref{fig:domain_shift}a shows a real depth image (pixel intensity corresponds to distance from camera) of the robotic gripper holding a peg and Figure~\ref{fig:domain_shift}b shows a simulated image of the same scene. Notice that the real image is noisier and there are several dropped pixels shown in white (e.g. the shadows behind the peg and inside the hole). If we run the policy learned in simulation directly on the real depth images, the system performs poorly because of these differences. This is known as the \emph{domain shift problem} and it must be addressed in some way in order to run experiments on the physical robot. A variety of techniques exist for mitigating this problem, e.g.~\cite{tobin2017domain} and~\cite{bousmalis2017using}. In these experiments, we used the pix2pix GAN approach~\cite{isola2017image} because it was a standalone piece of software that we could incorporate relatively easily. The pix2pix GAN learns a model that transforms a real image into an image that looks like what the simulator would have produced under similar circumstances. This is illustrated in Figure~\ref{fig:domain_shift}c. We trained the pix2pix GAN using 3200 paired real/simulated images from four different scenes. Then, we used it as a preprocessing step on real depth images. During testing, real depth images produced by the depth sensor were preprocessed by the pix2pix GAN and the result was input to the learned policy in order to select an action.

\begin{table}
\begin{center}
\begin{tabular}{ |c|c|c|c|c| } 
 \hline
& & \multicolumn{3}{|c|}{Failure Mode} \\
\hline
 Scene & Success & Collision & Not Recognize & Missed Hole \\ 
 & Rate &  & Hole &  \\ 
\hline
1 & 14/20 & 2 & 3 & 1 \\
2 & 19/20 & - & 1 & - \\ 
3 & 17/20 & 1 & 2 & - \\ 
4 & 18/20 & - & 1 & 1 \\ 
5 & 18/20 & - & - & 2 \\
\hline
\end{tabular}
\end{center}
\caption{Results from physical robot trials. Success rates and failure modes for each of five different scenes (see Figure~\ref{fig:exp_scenario}).}
\vspace{-1.0cm}
\label{table:physical_results}
\end{table}

\subsection{Experimental protocol and results}

We ran experimental trials on five different scenes with varying amounts of clutter. We ran 20 trials for each scene for a total of 100 trials. In each scene, clutter objects were selected and placed arbitrarily, but without blocking the hole (same protocol that was used in the simulated trials). All objects and placements were novel with respect to the scenes used for training. Each trial began with the manipulator in a randomly selected pose (not in collision) continued until either reaching the hole, colliding with an object or timing out. We used the same policy for all trials -- one that was trained in simulation using the full batch BQC cloning (pretraining) followed by TD learning (finetuning). Figure~\ref{fig:physical_trial} shows an example of one of our trials. Table~\ref{table:physical_results} shows the results of the 100 trials. Average success rates over all 100 trials was 86\% -- similar to the performance in simulation on the holdout testset (87\% for 4k scenes in Figure~\ref{fig:td_scenes}). 

\section{Conclusion} 
\label{sec:conclusion}

We explore \emph{planner cloning}, an approach that leverages the asymmetry in information that is available to the agent at train and test time. Since training is in simulation, full state information is available to the agent and it is possible to generate ``expert'' policy and value function rollouts from a full state planner (a graph based motion planner in our case). However, since only image observations are available to the agent at test time, it is necessary to project these plans onto a policy that the agent can execute. This happens via behavior cloning. This paper proposes a new behavior cloning method called Penalized $Q$ Cloning (PQC) that we demonstrate outperforms several algorithm ablations and baselines in simulation. Finally, we demonstrate that the resulting policies have similarly good performance for visual servoing tasks on a real robotic system.

\bibliographystyle{plainnat}
\bibliography{references}

\newpage

\end{document}